\documentclass[preprint,12pt]{elsarticle}

\usepackage{amssymb}
\usepackage{amsmath}
\usepackage{booktabs}
\usepackage{tabularx}
\usepackage{multirow}
\usepackage{array}
\usepackage{tabularx}
\usepackage{booktabs}
\usepackage{multirow}
\usepackage{adjustbox}

\newcolumntype{L}[1]{>{\raggedright\arraybackslash}p{#1}}
\newcolumntype{Y}{>{\raggedright\arraybackslash}X}

\journal{Optics \& Laser Technology}

\begin{document}

\begin{frontmatter}



\title{Self-Supervised Dual-Frequency Phase Decomposition for Single-Shot Composite Fringe Projection Profilometry}

\author[label1,label3]{Jin-Hyuk Seok}

\author[label2]{Yatong An\corref{cor1}}
\ead{yatong@meta.com}

\author[label1,label3]{Jae-Sang Hyun\corref{cor1}}
\ead{hyun.jaesang@yonsei.ac.kr}

\cortext[cor1]{Corresponding authors}

\affiliation[label1]{
    organization={Department of Mechanical Engineering, Yonsei University},
    city={Seoul},
    postcode={03722},
    country={South Korea}
}

\affiliation[label3]{
    organization={Yonsei Institute for Embodied Intelligence, Yonsei University},
    city={Seoul},
    postcode={03722},
    country={South Korea}
}

\affiliation[label2]{
    organization={Meta Reality Labs},
    city={Redmond},
    postcode={98052},
    state={Washington},
    country={USA}
}



\begin{abstract}

Single-shot fringe projection profilometry (FPP) has been actively studied for real-time measurement, dynamic object reconstruction, and motion-sensitive environments.
Composite fringe patterns are advantageous in single-shot FPP because multiple frequency components can be encoded in a single pattern, enabling phase ambiguity resolution.
Existing approaches mainly rely on Fourier transform-based methods or supervised deep learning methods.
However, Fourier transform-based methods often suffer from limited accuracy and degraded performance in complex regions, while supervised methods require dense phase or depth labels, which are costly to obtain.
In this work, we propose a self-supervised phase refinement framework for single-shot composite fringe patterns without requiring phase or depth labels.
The proposed method exploits the scale and direction relationships between low- and high-frequency phase gradients, improving the reliability of phase separation.
We also introduce a soft edge consistency loss to preserve object boundaries and fine geometric structures.
Experimental results show that the proposed method achieves MAE$_z$ and RMSE$_z$ of 0.367 mm and 1.804 mm, respectively, outperforming the best-performing transform-based baseline, which obtains 0.402 mm and 2.785 mm.
The proposed method also improves the valid-pixel ratio from 84.75\% to 95.07\%.
These results demonstrate the effectiveness of self-supervised dual-frequency phase refinement for reliable single-shot 3D reconstruction without ground-truth label supervision.

\end{abstract}

\begin{graphicalabstract}
  \includegraphics[width=\linewidth]{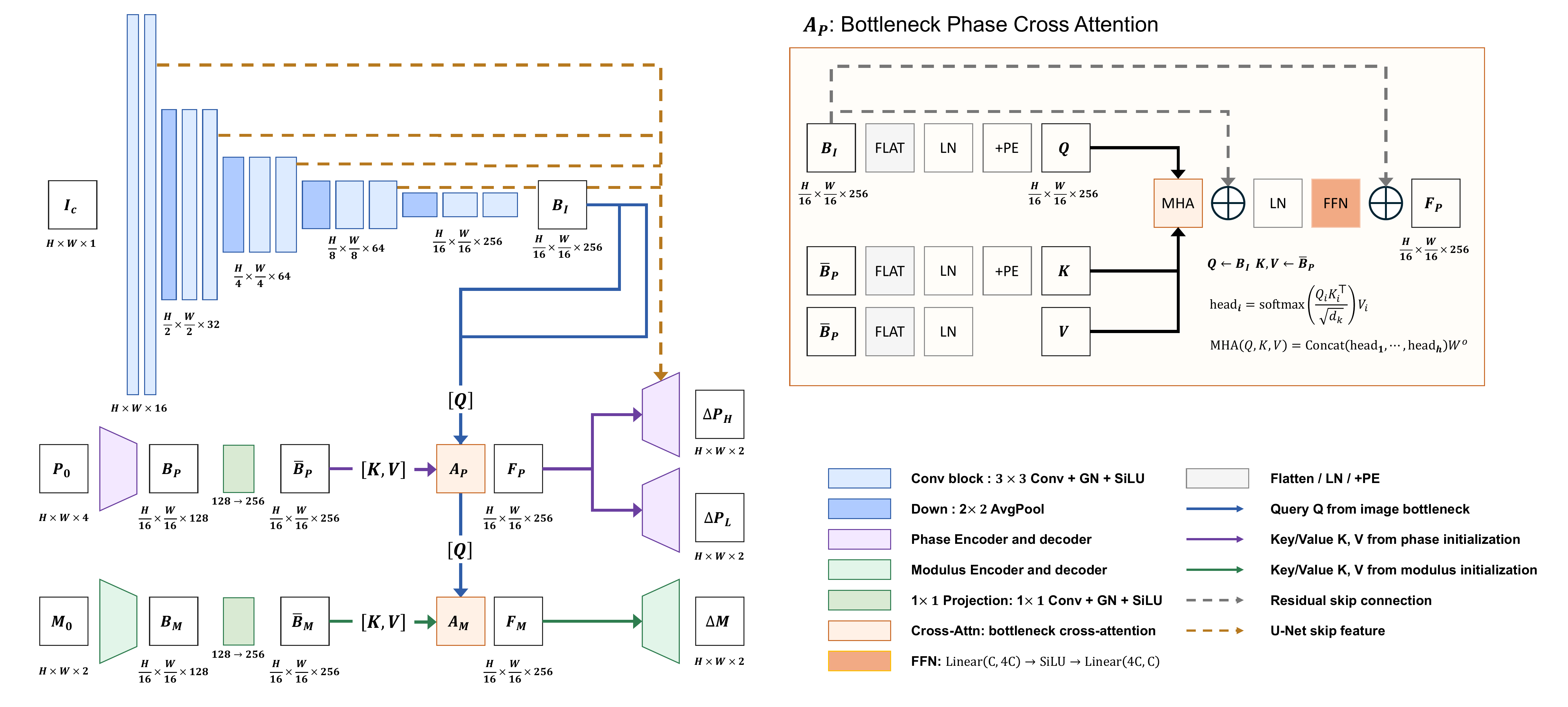}
\end{graphicalabstract}

\begin{highlights}

\item A self-supervised framework refines dual-frequency phases from one fringe image.
\item Training requires no ground-truth phase or depth labels.
\item Phase-gradient scale and direction consistency guides label-free refinement.
\item Soft edge consistency preserves object boundaries and fine geometric structures.
\item The method improves valid-pixel ratio and 3D depth accuracy over baselines.

\end{highlights}


\end{frontmatter}


\section{Introduction}

FPP is widely used as a non-contact, high-resolution, and high-accuracy 3D measurement technique~\cite{gorthi2010fringe, feng2018high, feng2021calibration}.
In a camera-projector system, the projector can be regarded as an inverse imaging device of the camera, and the projector coordinate corresponding to each camera pixel is determined by the absolute phase.
Conventional FPP systems sequentially project multiple fringe patterns, retrieve the wrapped phase using an N-step phase-shifting method, and obtain the absolute phase using additional patterns for Gray-code or multi-frequency phase unwrapping~\cite{zuo2018phase, zhang2018absolute}.
Although such multi-shot methods provide high accuracy, they suffer from low acquisition efficiency in dynamic object measurement.
Many studies have aimed to improve acquisition efficiency by accelerating projection and capture hardware, improving pattern encoding and coding strategies, and reducing the number of required patterns~\cite{zhang2010recent, zhang2018high}.
Recently, single-shot FPP has attracted increasing attention because it can estimate phase or depth from a single captured image, making it suitable for real-time measurement, dynamic object reconstruction, and motion-sensitive environments such as vibration or environmental disturbance~\cite{liu2024deep, wang2025single}.

Fourier transform profilometry (FTP) is one of the representative traditional approaches for single-shot phase retrieval~\cite{huang2010comparison}.
It recovers the phase by analyzing the global frequency spectrum of a captured fringe image and isolating the desired carrier component.
However, in practical measurements, surface depth variations, textures, and discontinuities can induce spatially varying fringe frequencies.
To handle such local frequency variations, windowed Fourier transform (WFT) performs localized spectral analysis using a spatial window~\cite{kemao2007two}, while wavelet transform (WT) provides a multi-resolution representation for localized frequency analysis~\cite{li2009spatial}.
Shearlet transform (ST) further exploits multi-scale and multi-directional representations, improving phase extraction in rapidly varying regions~\cite{gao2024one}.

Compared with learning-based methods, transform-based approaches are analytically interpretable and do not require training labels.
Nevertheless, their performance is highly dependent on manually selected parameters, including window size, mother wavelet, scale, direction, and threshold, and may require scene- or object-specific tuning.
This limitation becomes more pronounced in composite fringe patterns, where low- and high-frequency components coexist in a single image and can cause spectral overlap, frequency leakage, and component cross-talk, thereby degrading the accuracy of transform-based separation~\cite{li2022composite}.

Encoding additional phase cues into a single projected pattern is important for alleviating phase ambiguity in single-shot FPP.  A common strategy is to encode low- and high-frequency fringes into a single composite pattern~\cite{li2025deep, yue2007fourier}. The low-frequency phase provides a coarse cue for reducing fringe order ambiguity, whereas the high-frequency phase is advantageous for preserving fine geometric details~\cite{li2022composite, tang2024dual}. However, as discussed above, accurately separating different frequency components from a single composite image remains challenging, and transform-based methods are vulnerable to low- and high-frequency leakage.

To address the limitation, learning-based single-shot FPP methods have been actively researched. In particular, supervised learning-based methods directly predict phase or depth from a single fringe image and improve robustness~\cite{wang2023deep, wang2025end}. Recent studies have encoded coarse phase cues and fine phase details using dual-frequency or frequency-multiplexed composite patterns~\cite{li2022composite, li2025deep, chen2024deep, wu2024depth}, phase-shifting composite patterns~\cite{qin2025single}, spatially designed composite patterns~\cite{jiang2024deep}, or multi-channel color-coded composite patterns~\cite{fu2024deep}. However, these methods are generally based on fully supervised learning and require N-step phase shifting, Gray-code patterns, phase or depth labels, or labels generated from synthetic rendering~\cite{wang2021single}.

To reduce the dependence on ground-truth labels, label-efficient FPP methods have been investigated. Untrained and unsupervised approaches estimate phase or depth without phase or depth labels by exploiting physical or geometric constraints~\cite{yu2022untrained, yu2023untrained, fan2021unsupervised}. Yu et al. introduced untrained deep learning-based FPP methods that use geometric constraints in a stereo camera-projector system to alleviate the phase ambiguity of single sinusoidal fringes. By reprojecting the predicted unwrapped phase and depth into another camera view or the phase domain, these methods enforce consistency along epipolar geometry without pre-training~\cite{yu2022untrained, yu2023untrained}. Fan et al. proposed an unsupervised dual-frequency FPP framework, where the predicted depth is reprojected into the fringe domain and compared with captured dual fringe images of different frequencies~\cite{fan2021unsupervised}.

Weakly and semi-supervised FPP methods have also been explored to reduce the label cost of fully supervised learning ~\cite{gao2024weakly, tan2024weakly, wang2026end}. Gao et al. proposed a weakly supervised phase unwrapping method, where supervision is derived from high-frequency wrapped phase, one-period phase maps, and geometric constraints rather than accurate high-frequency absolute phase labels~\cite{gao2024weakly}. Tan et al. extended weak supervision to single-camera FPP depth estimation to reduce the dependence on dense depth ground truth~\cite{tan2024weakly}. Wang et al. introduced a semi-supervised end-to-end single-shot FPP framework that leverages both labeled and unlabeled data~\cite{wang2026end}. Although they also evaluated a self-supervised configuration, its performance was still limited compared with the semi-supervised setting. This indicates that self-supervised single-shot FPP remains challenging, especially when no direct phase or depth supervision is available.

Previous studies have proposed various strategies to reduce the dependence on ground-truth labels. 
However, self-supervised learning that directly exploits the inherent low- and high-frequency relationship within a single composite fringe image has not been sufficiently investigated. 
Existing label-free or self-supervised FPP methods are often based on single-frequency sinusoidal patterns or separately captured multi-frequency inputs. 
In contrast, a composite pattern contains both low-frequency cues and high-frequency details in a single image. 
Since these frequency components are modulated by the same object geometry, they share similar structural variations. 
Therefore, the gradient scale and direction relationship between low- and high-frequency phases can serve as an important self-supervised signal without requiring phase or depth labels.

In addition, self-supervised phase estimation may suffer from over-smoothing around object boundaries and depth discontinuities when it relies only on reconstruction consistency or smoothness constraints. 
Existing semi-supervised single-shot FPP used Sobel-filter-based edge-aware phase smoothing to alleviate this issue~\cite{wang2026end}, and edge-aware smoothness losses have also been widely used in self-supervised depth estimation in other research fields~\cite{chen2023self}. 
Motivated by these works, we observe that phase, modulation amplitude, and texture share common structural discontinuities at object boundaries. 
Based on this observation, we define soft edge maps from second-order phase variation, modulation amplitude, and texture, and enforce consistency among them to alleviate boundary over-smoothing.

The proposed method takes a single composite fringe image as input and estimates low- and high-frequency wrapped phases together with their corresponding modulation amplitudes. 
We reconstruct the input image to form an image-intensity reconstruction loss, and further introduce low- and high-frequency phase gradient consistency and phase-modulation-texture soft edge consistency losses. 
As a result, the proposed method performs composite phase decomposition using only self-supervised constraints derived from a single composite image, without auxiliary weak labels or ground-truth phase or depth labels.

The main contributions of this research are summarized as follows:

\begin{itemize}
    \item \textbf{Self-supervised composite phase decomposition.}
    We propose a self-supervised single-shot framework that decomposes a captured composite fringe image into low- and high-frequency phase components without requiring ground-truth phase or depth labels.

    \item \textbf{Gradient-structural regularization between dual-frequency phases.}
    We introduce structural constraints based on the scale and direction relationships between the gradients of low- and high-frequency phases. 
    By exploiting the shared geometric variation of the two frequency components, the proposed regularization improves the reliability of phase separation from a single composite pattern.

    \item \textbf{Soft edge consistency for boundary preservation.}
    We introduce a soft edge consistency loss that aligns structural edges among phase variations, modulation amplitudes, and texture. 
    This loss mitigates over-smoothing and helps preserve object boundaries and fine geometric structures during self-supervised phase refinement.
\end{itemize}

\begin{table}[!htbp]
\centering
\footnotesize
\setlength{\tabcolsep}{4pt}
\renewcommand{\arraystretch}{1.15}

\begin{tabularx}{\linewidth}{L{3.1cm} Y Y Y}
\toprule
 &
\textbf{Transform-based} &
\textbf{Learning-based (Supervised)} &
\textbf{Proposed Method (Self-supervised)} \\
\midrule

Input &
Single composite pattern &
Single composite pattern &
Single composite pattern \\

GT supervision &
Not required &
Required &
Not required \\

Learning requirement &
Not required &
Required &
Required \\
\midrule
Inference tuning &
Required; window/scale/threshold &
Not required after training &
Not required after training \\

Init./tuning sensitivity &
High; suboptimal settings directly degrade results &
Less relevant (feed-forward after training) &
Low; physics-guided optimization refines suboptimal initialization \\
\midrule
Accuracy &
Lower &
Higher &
Medium-High \\

Model-based \& physics grounding &
High; analytical, frequency-domain model-based &
Low; data-driven, limited physical transparency &
Medium--High; physics-guided refinement with model-based initialization \\

\midrule
Main advantage &
Analytical and physically grounded; no training data &
Fast and accurate with sufficient GT &
No GT phase; improved accuracy over WFT while retaining physics-based reliability \\

Main limitation &
Lower accuracy; parameter sensitivity &
Requires GT; limited interpretability &
Typically below fully supervised accuracy \\
\bottomrule
\end{tabularx}

\caption{Comparison of WFT-based conventional, supervised learning-based, and the proposed physics-guided refinement approach. All methods take a single composite pattern as input.}
\label{tab:method_comparison}
\end{table}

\section{Related Works}

\subsection{Windowed Fourier transform}

In this work, the proposed method adopts the WFT output as an initial phase estimate. The WFT of an input image ${f(x, y)}$ can be expressed as Eq.~\eqref{eq:window fourier transform}. ${g_{u,v,\zeta,\eta}(x,y)}$ denotes a local basis constructed as the product of a Gaussian window function and a Fourier basis as shown in Eq.~\eqref{eq:basis}. After applying band-pass filtering under an appropriate frequency range in the WFT domain, we obtain the reconstructed image in complex form and wrapped phase as Eq.~\eqref{eq:filtered inverse} and Eq.~\eqref{eq:phase}.

\begin{equation}
Sf(u,v,\zeta,\eta)=\int_{-\infty}^{\infty}\int_{-\infty}^{\infty}f(x, y) g^*_{u,v,\zeta,\eta}(x,y)\, dx\, dy
\label{eq:window fourier transform}
\end{equation}

\begin{equation}
g_{u,v,\zeta,\eta}(x,y)=g(x-u,y-v)\exp(j \zeta x+j\eta y)
\label{eq:basis}
\end{equation}

\begin{equation}
\overline{f}(x,y)={{1}\over{4\pi^2}}\int_{-\infty}^{\infty}\int_{-\infty}^{\infty}\int_{-\eta_l}^{\eta_h}\int_{-\zeta_l}^{\zeta_h}\overline{Sf}(u,v,\zeta,\eta) g_{u,v,\zeta,\eta}(x,y)\,d\zeta\,d\eta\,du\, dv
\label{eq:filtered inverse}
\end{equation}

\begin{equation}
\phi(x,y) = \tan^{-1}\!\left( 
    \frac{\operatorname{Im}[\overline{f}(x,y)]}{\operatorname{Re}[\overline{f}(x,y)]}
\right)
\label{eq:phase}
\end{equation}

\subsection{Phase-based Candidate Reduction for Stereo Matching}

We adopt phase-based stereo matching for 3D reconstruction. 
Previous stereo camera-projector methods resolve phase ambiguity without additional unwrapping patterns by using wrapped phase consistency and geometric constraints~\cite{fu2024high,tang2024dual}. 
In particular, Fu et al.~\cite{fu2024high} used stereo intensity and phase information for candidate reduction, while Tang et al.~\cite{tang2024dual} showed that low-frequency wrapped phase reduces ambiguity in composite-pattern matching. 
Following this idea, we define the correspondence candidates as
\begin{equation}
\mathcal{C}(\mathbf{x}_L)
=
\left\{
\mathbf{x}_R
\ \middle|\
d\!\left(\phi_l^L(\mathbf{x}_L), \phi_l^R(\mathbf{x}_R)\right) < \tau_l,\ 
d\!\left(\phi_h^L(\mathbf{x}_L), \phi_h^R(\mathbf{x}_R)\right) < \tau_h
\right\},
\end{equation}
where 
$d(a,b)=|\operatorname{atan2}(\sin(a-b),\cos(a-b))|$.

The final correspondence is then selected from $\mathcal{C}(\mathbf{x}_L)$ by favoring candidates with high image intensity similarity and low wrapped phase discrepancy.

\section{Principles}
\label{Principle}
\begin{figure}[!htbp]
  \centering
  \includegraphics[width=\linewidth]{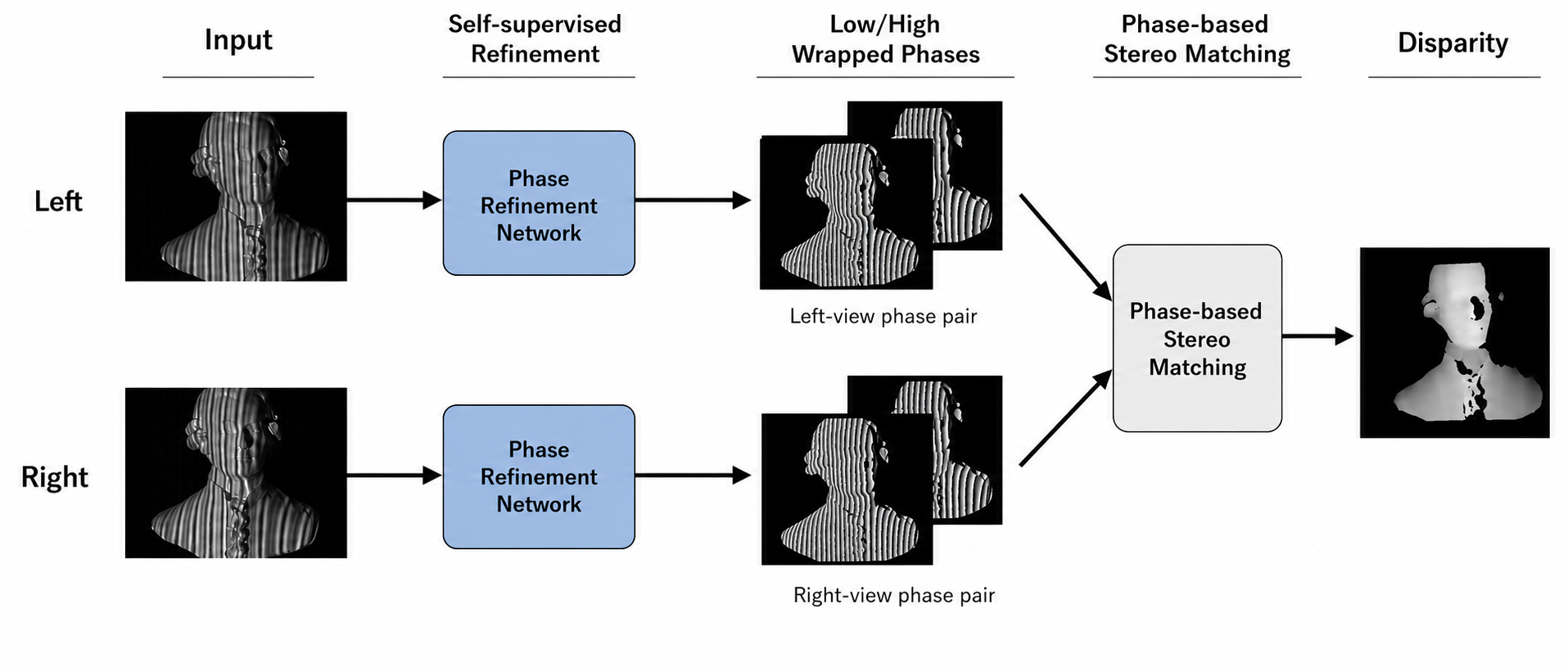}
  \caption{Overview of the proposed single-shot structured-light reconstruction framework. The predicted low- and high-frequency wrapped phases are used for stereo correspondence, and the final 3D coordinates are recovered by triangulation.}
  \label{fig:recon_arch}
\end{figure}

The proposed framework starts from a single dual-frequency composite fringe image, where low- and high-frequency sinusoidal components are modulated in one exposure.
We first apply the Windowed Fourier Transform (WFT) to obtain initial estimates of the low- and high-frequency components.
Since the WFT initialization is physically meaningful but locally imperfect due to spectral leakage, texture interference, and spatially varying fringe frequency, we treat it as an initial representation to be refined rather than as a final prediction.

To this end, we decompose the WFT outputs into unit phase representations and modulus maps, and refine them using an image-conditioned residual network.
The network separately encodes the composite image, the phase initialization, and the modulus initialization, and fuses them through bottleneck cross-attention.
The refined low- and high-frequency phase components are then optimized using self-supervised losses based on image reconstruction and the structural consistency between the two frequency components. After phase refinement, stereo correspondence is determined using the predicted low- and high-frequency wrapped phases, and 3D coordinates are recovered by triangulation as shown in Fig.~\ref{fig:recon_arch}.

\subsection{Pattern Design}
\label{subsec1}

\begin{figure}[!htbp]
  \centering
  \includegraphics[width=\linewidth]{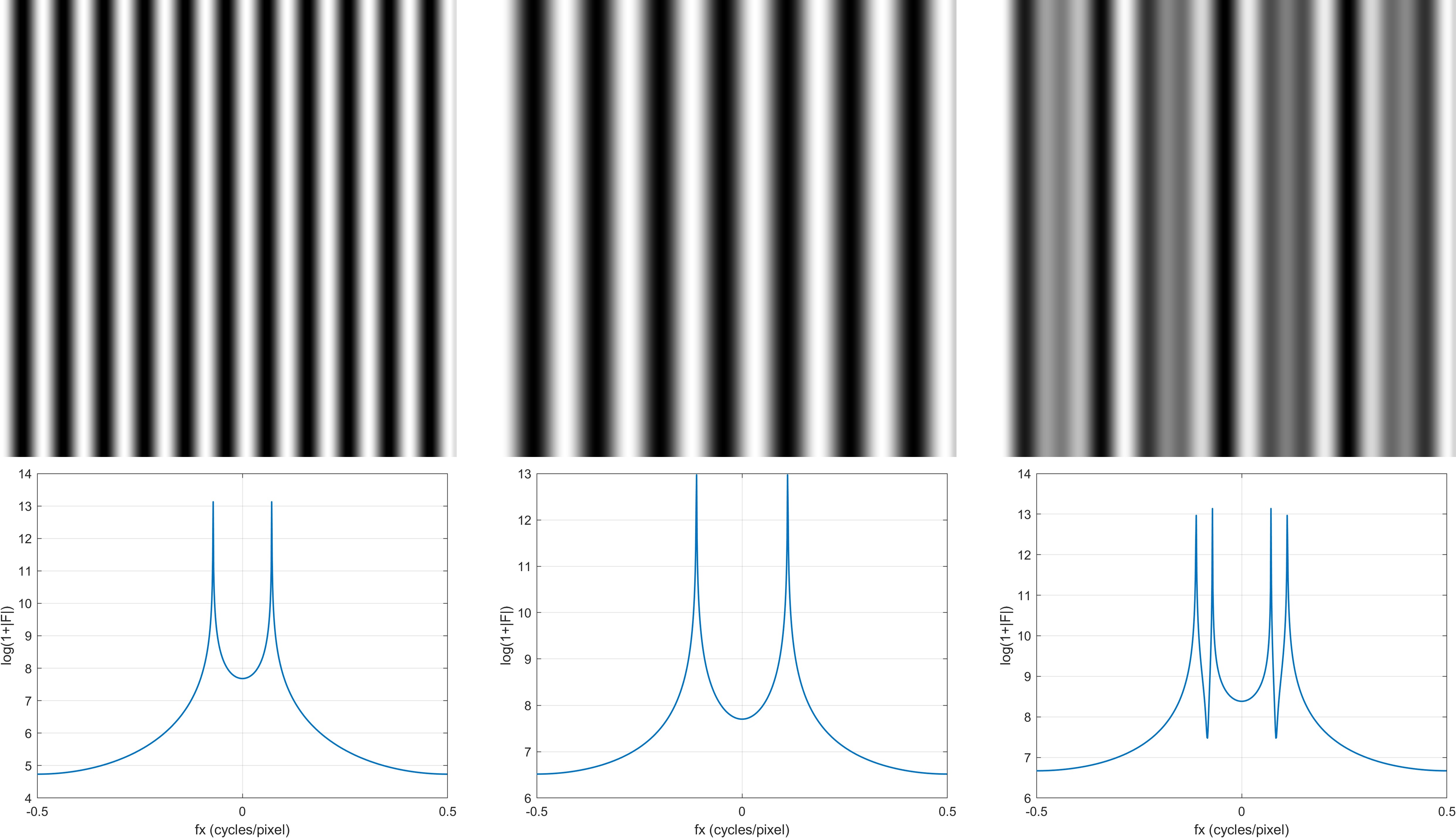}
  \caption{Top row shows the high-frequency, low-frequency, and composite fringe patterns; bottom row shows the respective DC-centered 1D Fourier spectra along the horizontal axis.}
  \label{fig:composite_design}
\end{figure}

\begin{equation}
I_{cp}^P(x,y) = A + B(\cos({{2\pi } \over {\lambda_H}}x) + \cos({{2\pi } \over {\lambda_L}}x))
\label{eq:composite}
\end{equation}

The composite fringe pattern used in this work is defined by Eq.~\eqref{eq:composite}. It is constructed as the sum of a high-frequency and a low-frequency fringe pattern, and is normalized to the intensity range of $[0, 255]$ using constant $A$ and $B$. The wavelengths are denoted by $\lambda_h$ and $\lambda_l$, and we set $\lambda_h = 9$ and $\lambda_l = 14$. Obtaining an absolute phase which fully covers the projector width with multi-frequency phase unwrapping requires either increasing the number of frequency components or choosing wavelengths that are sufficiently close to each other. However, incorporating too many frequency components degrades the SNR of the desired signal and makes it harder to separate it from noise. Selecting closely spaced frequencies can also lead to spectral overlap and frequency leakage. Using only a single-frequency pattern is also undesirable. The phase ambiguity becomes more severe during the stereo matching due to the lack of clues. Considering those trade-offs, we adopt a dual-frequency composite pattern. Although the design does not provide a globally covering absolute phase, it yields higher SNR for the desired components and enables more robust correspondence estimation than a single-frequency configuration. As shown in Fig.~\ref{fig:composite_design}, the pattern is designed to be well separated in the Fourier domain, with a sufficient band to minimize spectral interference. 


\begin{figure}[!htbp]
  \centering
  \includegraphics[width=\linewidth]{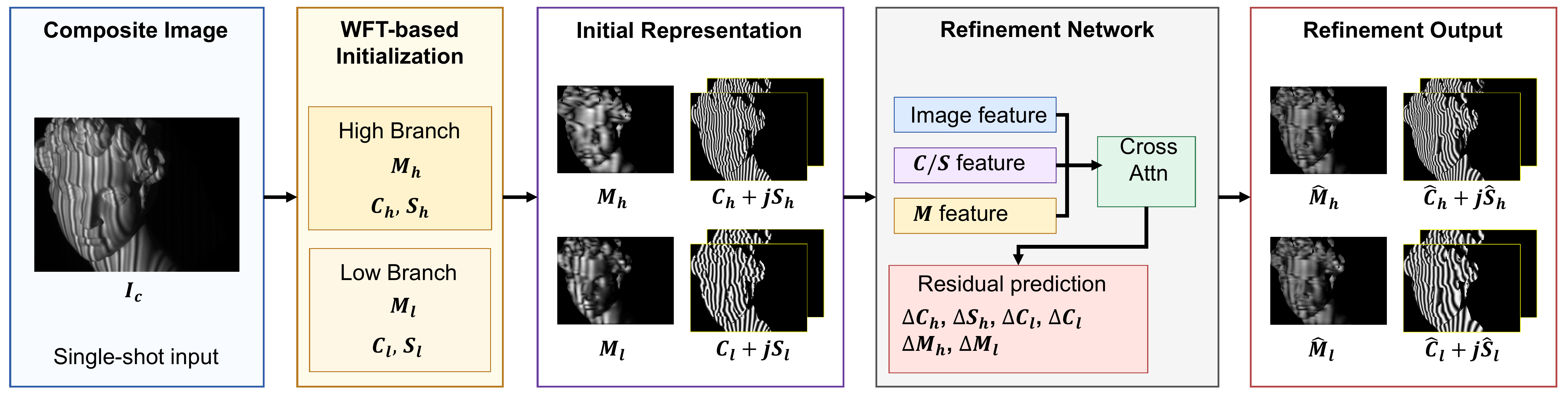}
  \caption{Overview of the proposed phase refinement process. The composite image is first decomposed by WFT into initial low- and high-frequency representations, which are then refined by the proposed residual network.}
  \label{fig:refine_overview}
\end{figure}

\subsection{Network Architecture}

The refinement process from the composite image to the WFT initialization and the final refined phase representations is summarized in Fig.~\ref{fig:refine_overview}. The WFT provides a physically meaningful initial estimate, but it can be locally inaccurate due to spectral leakage, texture interference, and spatially varying fringe frequency. Therefore, we formulate the network as a residual refinement module rather than predicting the phase representation from scratch under self-supervision. We decouple the WFT response into a unit phase representation and a modulus map. For each frequency component $n\in\{H,L\}$, the WFT-derived raw cosine-sine components are denoted as $(C_n,S_n)$.
The modulus and unit phase representations are computed as
\begin{equation}
\tilde{M}_n=\sqrt{C_n^2+S_n^2+\epsilon},
\label{eq:modulus_decomposition}
\end{equation}
\begin{equation}
(\tilde{C}_n,\tilde{S}_n)=\frac{(C_n,S_n)}{\tilde{M}_n}.
\label{eq:unit_phase_decomposition}
\end{equation}
We then define the initial phase and modulus representations as
\begin{equation}
\mathbf{P}_0=(\tilde{C}_H,\tilde{S}_H,\tilde{C}_L,\tilde{S}_L),
\qquad
\mathbf{M}_0=(\tilde{M}_H,\tilde{M}_L).
\label{eq:initial_representations}
\end{equation}

\begin{figure}[!htbp]
  \centering
  \includegraphics[width=\linewidth]{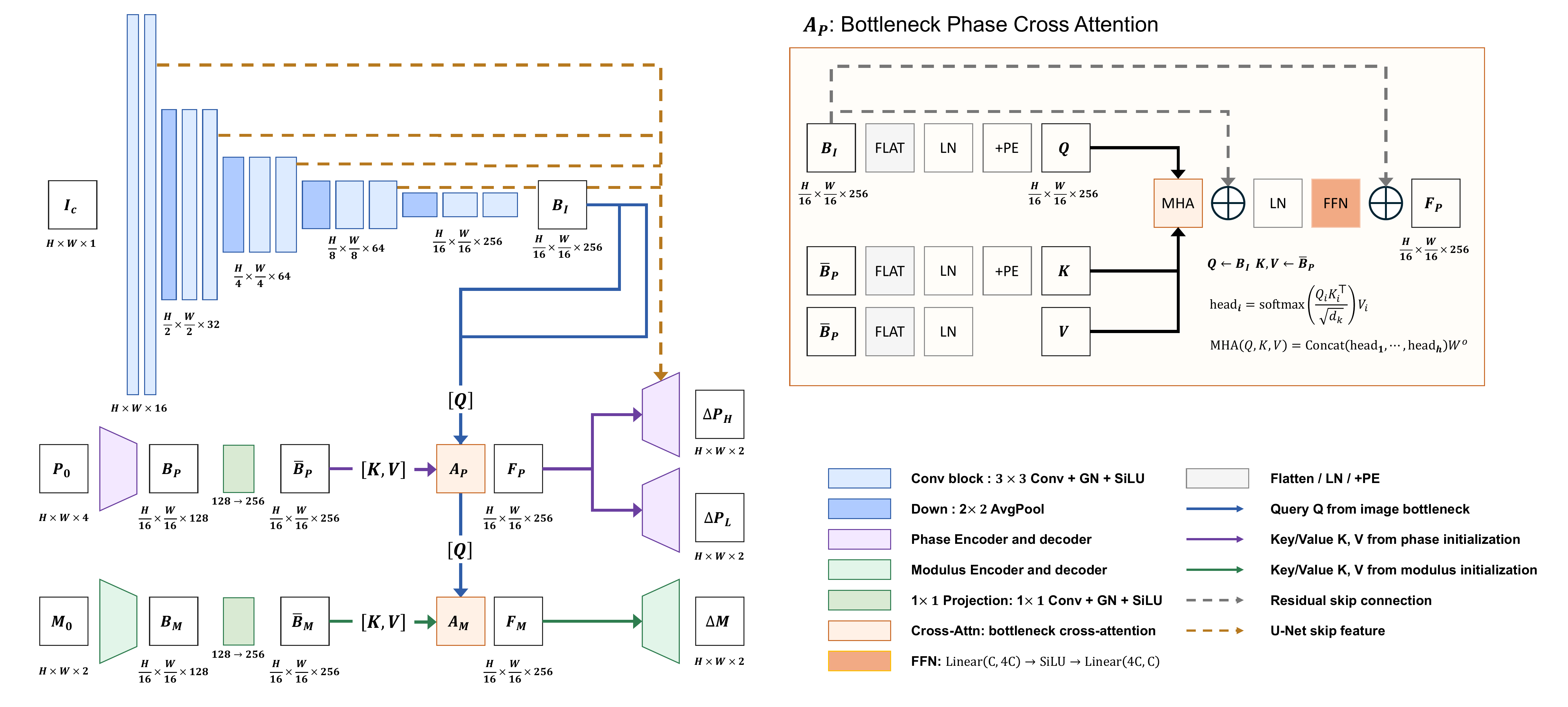}
  \caption{Detailed architecture of the proposed unsupervised phase refinement module. The composite image, phase initialization, and modulus initialization are encoded separately and fused through bottleneck cross-attention for residual refinement.}
  \label{fig:uprm_module}
\end{figure}

As illustrated in Fig.~\ref{fig:uprm_module}, the composite image $I_c$, phase initialization $\mathbf{P}_0$, and modulus initialization $\mathbf{M}_0$ are processed by three U-Net encoders.
The image encoder extracts the observation bottleneck feature $B_I$, while the phase and modulus encoders extract $B_P$ and $B_M$ from the WFT-initialized representations.

\begin{equation}
B_I=\mathcal{E}_I(I_c),\qquad
B_P=\mathcal{E}_P(\mathbf{P}_0),\qquad
B_M=\mathcal{E}_M(\mathbf{M}_0),
\label{eq:encoder_features}
\end{equation}
where \(\mathcal{E}_I\), \(\mathcal{E}_P\), and \(\mathcal{E}_M\) denote the image, phase-initialization, and modulus-initialization encoders, respectively.

Before bottleneck fusion, $B_P$ and $B_M$ are projected to the same channel dimension as $B_I$:
\begin{equation}
\bar{B}_P=\psi_P(B_P),\qquad
\bar{B}_M=\psi_M(B_M),
\label{eq:bottleneck_projection}
\end{equation}
where $\psi_P$ and $\psi_M$ denote $1\times1$ projection layers.

\begin{table}[!htbp]
\centering
\setlength{\tabcolsep}{5pt}
\renewcommand{\arraystretch}{1.12}
\begin{tabularx}{\linewidth}{lX}
\hline
Notation & Description \\
\hline
$I_c$ & Input composite fringe image. \\
$n\in\{H,L\}$ & Frequency index for high- and low-frequency components. \\
$(C_n,S_n)$ & WFT-derived raw cosine-sine components for frequency $n$. \\
$\tilde{(\cdot)}$ & Normalized initial component derived from the WFT response. \\
$\hat{(\cdot)}$ & Refined output from the proposed network. \\
$(\tilde{C}_n,\tilde{S}_n),\tilde{M}_n$ & Initial unit phase and modulus representations for frequency $n$. \\
$\mathbf{P}_0,\mathbf{M}_0$ & Initial phase and modulus representations used as network inputs. \\
$B_I$ & Bottleneck feature from the composite image stream. \\
$B_P,B_M$ & Bottleneck features from the phase and modulus initialization streams. \\
$\bar{B}_P,\bar{B}_M$ & Channel-projected guidance features. \\
$F_P,F_M$ & Attention-fused phase and modulus features. \\
$\Delta P_n,\Delta\mathbf{M}$ & Predicted phase and modulus residuals. \\
\hline
\end{tabularx}
\caption{Notation used in the proposed refinement network.}
\label{tab:notation}
\end{table}

After channel alignment, the bottleneck features are fused using bottleneck cross-attention modules. The image bottleneck feature is used as the query, while the projected phase or modulus guidance feature is used as the key and value. Since the image feature provides local observation context and the WFT initialization serves as both a physical prior and a refinement target, the attention module allows the network to determine which parts of the initialization should be preserved or corrected.

For each branch $r\in\{P,M\}$, let $\bar{B}_r$ denote the projected guidance feature.
The image bottleneck feature is used as the query, while $\bar{B}_r$ is used as the key and value:
\begin{equation}
Q=\operatorname{LN}(\operatorname{Flatten}(B_I))+\lambda\mathrm{PE},
\label{eq:attn_query}
\end{equation}
\begin{equation}
K_r=\operatorname{LN}(\operatorname{Flatten}(\bar{B}_r))+\lambda\mathrm{PE},\qquad
V_r=\operatorname{LN}(\operatorname{Flatten}(\bar{B}_r)).
\label{eq:attn_key_value}
\end{equation}

The fused phase and modulus features are obtained as
\begin{equation}
F_P=\mathcal{A}_P(B_I,\bar{B}_P),\qquad
F_M=\mathcal{A}_M(B_I,\bar{B}_M),
\label{eq:attention_fusion}
\end{equation}
where \(\mathcal{A}_P\) and \(\mathcal{A}_M\) denote the phase and modulus bottleneck cross-attention modules, respectively.

The attention-fused phase feature \(F_P\) is fed into two separate decoder heads to predict the high- and low-frequency phase residuals. The modulus-fused feature \(F_M\) is decoded by another decoder head to predict the modulus residual. Each decoder uses U-Net-style skip features from the image encoder to recover full-resolution local structures.

The decoder outputs are defined as
\begin{equation}
\Delta P_H=\mathcal{D}_H(F_P),\qquad
\Delta P_L=\mathcal{D}_L(F_P),\qquad
\Delta\mathbf{M}=\mathcal{D}_M(F_M).
\label{eq:decoder_residuals}
\end{equation}

The predicted phase residuals are added to the initial unit phase representations and then projected back onto the unit circle to enforce the cosine-sine unit-norm constraint.

For $n\in\{H,L\}$, let $\Delta P_n=(\Delta C_n,\Delta S_n)$.
The residual update is first computed as

\begin{equation}
(C_n',S_n')=(\tilde{C}_n,\tilde{S}_n)+\Delta P_n,\qquad n\in\{H,L\}.
\label{eq:phase_residual_update}
\end{equation}

The updated pair is then projected back onto the unit circle:

\begin{equation}
(\hat{C}_n,\hat{S}_n)
=
\frac{(C_n',S_n')}
{\sqrt{(C_n')^2+(S_n')^2+\epsilon}}.
\label{eq:unit_projection}
\end{equation}

The modulus residual is added to the initial modulus representation and passed through a sigmoid function.
The modulus representation is refined as
\begin{equation}
\hat{\mathbf{M}}=\sigma(\mathbf{M}_0+\Delta\mathbf{M}),
\label{eq:modulus_update}
\end{equation}
where $\sigma(\cdot)$ denotes the sigmoid function.

\subsection{Loss Function}

The captured composite image is modeled as the sum of a coarse texture/background component and two modulated sinusoidal components:
\begin{equation}
I_c
\approx
T
+
\sum_{n\in\{H,L\}}
M_n C_n,
\qquad C_n=\cos\phi_n .
\label{eq:obs_model_dual}
\end{equation}
Based on this observation model, the proposed objective consists of an image reconstruction loss, a phase consistency loss, a modulation regularization loss, and a shared edge consistency loss.
The reconstruction loss enforces consistency with the captured composite image, while the remaining terms provide self-supervised priors for phase and modulation refinement.

\subsubsection{Image reconstruction loss}

Based on the observation model in Eq.~\eqref{eq:obs_model_dual}, the composite fringe image should be reconstructed from the coarse texture component and the predicted low- and high-frequency sinusoidal components.
We use $\tilde{T}$ as a coarse texture estimate obtained by removing the extracted fringe components using WFT.
The reconstructed composite image is defined as
\begin{equation}
\hat{I}_c
=
\tilde{T}
+
\sum_{n\in\{H,L\}}
\hat{M}_n\hat{C}_n .
\label{eq:image_reconstruction}
\end{equation}

The reconstruction loss is defined using an L1 intensity term and an SSIM term:
\begin{equation}
\mathcal{L}_{\mathrm{rec}}
=
\left|
\hat{I}_c-I_c
\right|
+
\lambda_{\mathrm{SSIM}}
\left(
1-\mathrm{SSIM}(\hat{I}_c,I_c)
\right).
\label{eq:loss_rec}
\end{equation}

\subsubsection{Phase loss}

The phase loss consists of two terms based on the properties of phase gradients. A magnitude consistency term, which enforces that the gradient magnitudes of the high- and low-frequency phases are inversely proportional to their wavelengths, and a direction consistency term, which encourages the two gradient vectors to have the same orientation.

Although the phase can be computed via the arctangent function, directly differentiating the wrapped phase leads to error due to its 2$\pi$ periodic discontinuities. To avoid this issue, we compute phase gradients using only the observed cosine and sine components as shown in Eq.~\eqref{eq:unit_projection}. Then the phase gradient vector is computed as shown in Eq.~\eqref{eq:phase_grad_from_sincos_hat}. The magnitude of the gradient of the phase is expressed as shown in Eq.~\eqref{eq:phase_grad_mag_hat}.

\begin{equation}
\nabla \hat{\phi}_n(x,y)
=
\frac{
\hat{C}_n(x,y)\,\nabla \hat{S}_n(x,y)
-
\hat{S}_n(x,y)\,\nabla \hat{C}_n(x,y)
}{
\hat{C}_n(x,y)^2+\hat{S}_n(x,y)^2+\varepsilon
},
\qquad n\in\{h,l\}.
\label{eq:phase_grad_from_sincos_hat}
\end{equation}

\begin{equation}
\hat{m}_n(x,y)=\left\lVert \nabla \hat{\phi}_n(x,y) \right\rVert_2
=
\sqrt{
\big(\partial_x \hat{\phi}_n(x,y)\big)^2
+
\big(\partial_y \hat{\phi}_n(x,y)\big)^2
+\varepsilon
}.
\label{eq:phase_grad_mag_hat}
\end{equation}

The scale-consistency loss, $\mathcal{L}_{\mathrm{scale}}$, enforces that the gradient magnitudes of the high- and low-frequency phases are inversely proportional to their frequencies, as shown in Eq.~\eqref{eq:loss_scale}. We compare the magnitudes in the log domain because it is less sensitive than direct subtraction, which helps reduce overfitting and encourages smoother phase-gradient fields.

\begin{equation}
\mathcal{L}_{\mathrm{scale}}
=
\left\langle
\left|
\log\!\big(1+\hat{m}_h\big)
-
\log\!\big(1+r\,\hat{m}_l\big)
\right|
\right\rangle ,
\qquad
r=\frac{\lambda_h}{\lambda_l}.
\label{eq:loss_scale}
\end{equation}

$\mathcal{L}_{\mathrm{dir}}$ is motivated by the observation that both frequency components are modulated by the same surface geometry and thus share similar local orientation changes. Accordingly, the loss is designed to use the cosine similarity between the two gradient vectors, as formulated in Eq.~\eqref{eq:loss_dir}, to align the gradient directions of the high- and low-frequency phases.

\begin{equation}
\mathcal{L}_{\mathrm{dir}}
=
\left\langle
1-
\frac{
\nabla \hat{\phi}_h \cdot \nabla \hat{\phi}_l
}{
\left\lVert \nabla \hat{\phi}_h \right\rVert_2\,
\left\lVert \nabla \hat{\phi}_l \right\rVert_2+\varepsilon
}
\right\rangle .
\label{eq:loss_dir}
\end{equation}

$\mathcal{L}_{\mathrm{cont}}$ is implemented as a total-variation (TV) regularizer, defined in Eq.~\eqref{eq:charbonnier_tv} and Eq.~\eqref{eq:loss_cont}. This regularization term suppresses undesired spatial fluctuations in the phase-gradient magnitude, eliminating staircase and noise artifacts.

\begin{equation}
\mathrm{TV}_{\mathrm{Ch}}(u)
=
\left\langle
\sqrt{(\partial_x u)^2+\epsilon_{\mathrm{tv}}^{\,2}}
+
\sqrt{(\partial_y u)^2+\epsilon_{\mathrm{tv}}^{\,2}}
\right\rangle .
\label{eq:charbonnier_tv}
\end{equation}

\begin{equation}
\mathcal{L}_{\mathrm{cont}}
=
\mathrm{TV}_{\mathrm{Ch}}(\hat{m}_h)
+
\mathrm{TV}_{\mathrm{Ch}}(\hat{m}_l).
\label{eq:loss_cont}
\end{equation}

Finally, the overall phase loss is defined as a weighted sum of the three terms as shown in Eq.~\eqref{eq:phase_loss}.

\begin{equation}
\mathcal{L}_\mathrm{phase}
=
\lambda_{\mathrm{dir}}\mathcal{L}_{\mathrm{dir}}
+
\lambda_{\mathrm{scale}}\mathcal{L}_{\mathrm{scale}}
+
\lambda_{\mathrm{cont}}\mathcal{L}_{\mathrm{cont}}.
\label{eq:phase_loss}
\end{equation}

\subsubsection{Modulation loss}

The high- and low-frequency modulation amplitudes are affected by common factors such as surface reflectance, illumination strength, and camera/projector response. 
Therefore, although their absolute values may vary spatially, their relative ratio is expected to change smoothly within locally consistent regions. 
To suppress noise-like oscillations and staircasing artifacts in the predicted modulus maps, we regularize the spatial variation of the modulation ratio using a Charbonnier-TV loss:
\begin{equation}
\mathcal{L}_{\mathrm{mod}}^{\mathrm{ratio}}
=
\mathrm{TV}_{\mathrm{Ch}}\!\left(
\frac{\hat{M}_H}{\hat{M}_L+\epsilon}
\right).
\label{eq:loss_mod_sigma_ratio}
\end{equation}
This relaxed constraint encourages a piecewise-smooth ratio field without enforcing a globally constant modulation ratio.

\subsubsection{Shared edge consistency loss}

In locally continuous regions, the phase gradient should vary smoothly, whereas structural boundaries can induce abrupt changes in the phase gradient. 
Since such boundaries are often reflected in both texture and modulation amplitude, we encourage their edge structures to be spatially consistent. 
For each frequency component $n\in\{H,L\}$, we construct soft edge maps:
\begin{equation}
\begin{aligned}
E_{\phi_n}
&= \sigma\left(
\alpha_\phi
\left(
\left\|\nabla\|\nabla\phi_n\|\right\|
-\tau_\phi
\right)
\right), \\
E_{M_n}
&= \sigma\left(
\alpha_M
\left(
\|\nabla\log(M_n+\epsilon)\|
-\tau_M
\right)
\right), \\
E_T
&= \sigma\left(
\alpha_T
\left(
\|\nabla T_{\mathrm{WFT}}\|
-\tau_T
\right)
\right).
\end{aligned}
\label{eq:soft_edge_maps}
\end{equation}
Here, $T_{\mathrm{WFT}}$ is a coarse texture estimate obtained after removing the extracted fringe components.
The edge consistency loss is defined as
\begin{equation}
\mathcal{L}_{\mathrm{edge}}
=
\frac{1}{2}
\sum_{n\in\{H,L\}}
\left(
\|E_{\phi_n}-E_T\|_1
+
\|E_{M_n}-E_T\|_1
\right).
\label{eq:loss_edge}
\end{equation}

\section{Experiments}

\subsection{Experimental Setup}

\begin{figure}[!htbp]
    \centering
    \includegraphics[width=0.5\linewidth]{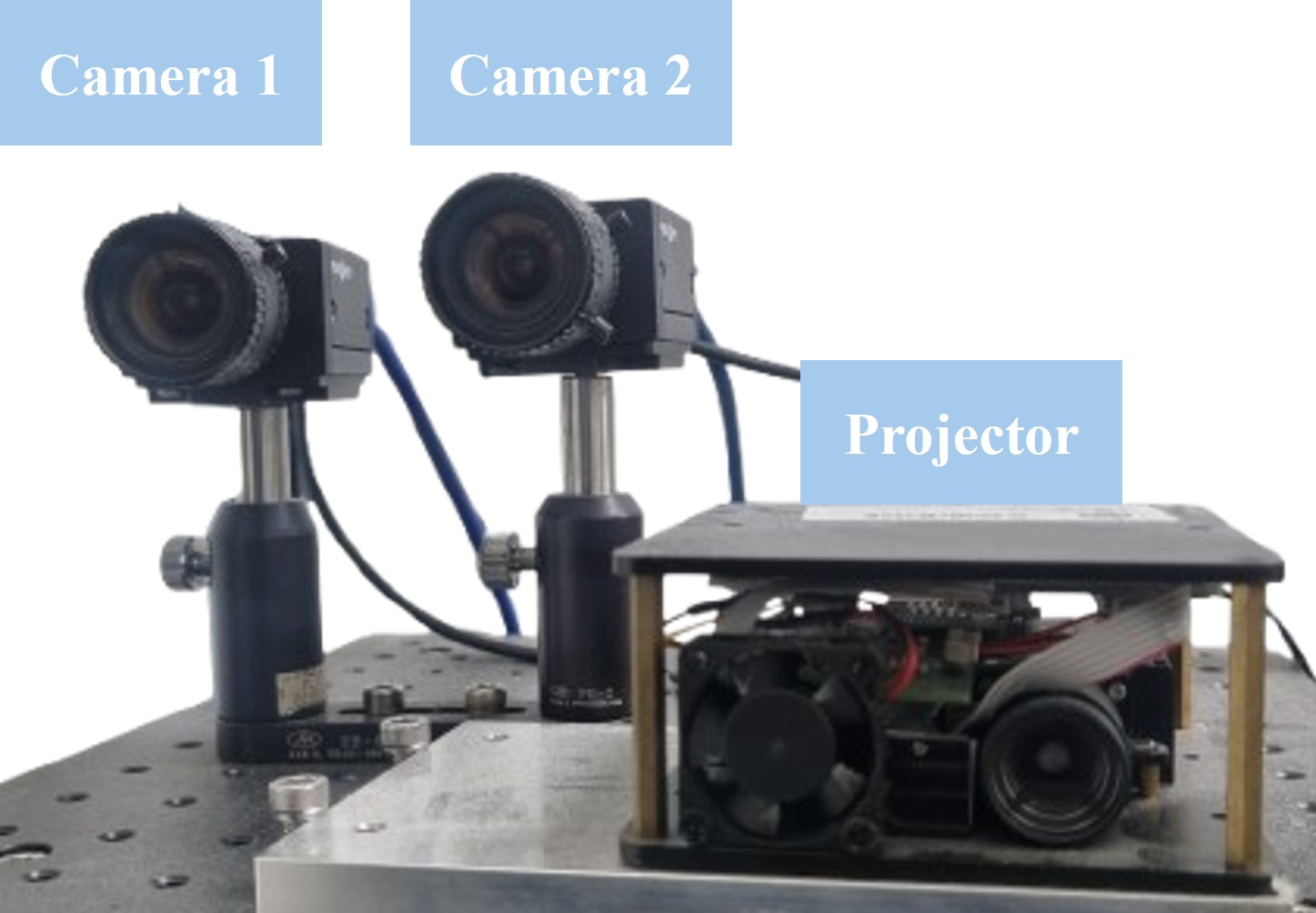}
    \caption{Experimental setup.}
    \label{fig:exp_setup}
\end{figure}

The experimental setup is shown in Fig.~\ref{fig:exp_setup}. The system employs two CCD cameras (FLIR Grasshopper3 GS3-U3) with 8 mm lenses at 640${\times}$480 resolution. Stereo camera calibration is performed with Zhang's method~\cite{zhang2006novel}. A Texas Instruments LightCrafter 4500 DLP projector is integrated, with 912 ${\times}$ 1140 projection resolution. In this setup, projector calibration is not required for our unsupervised phase retrieval pipeline but required only for evaluation. The composite pattern was designed using two spatial periods of 9 and 14 pixels.

\begin{figure}[t]
    \centering
    \includegraphics[width=\linewidth]{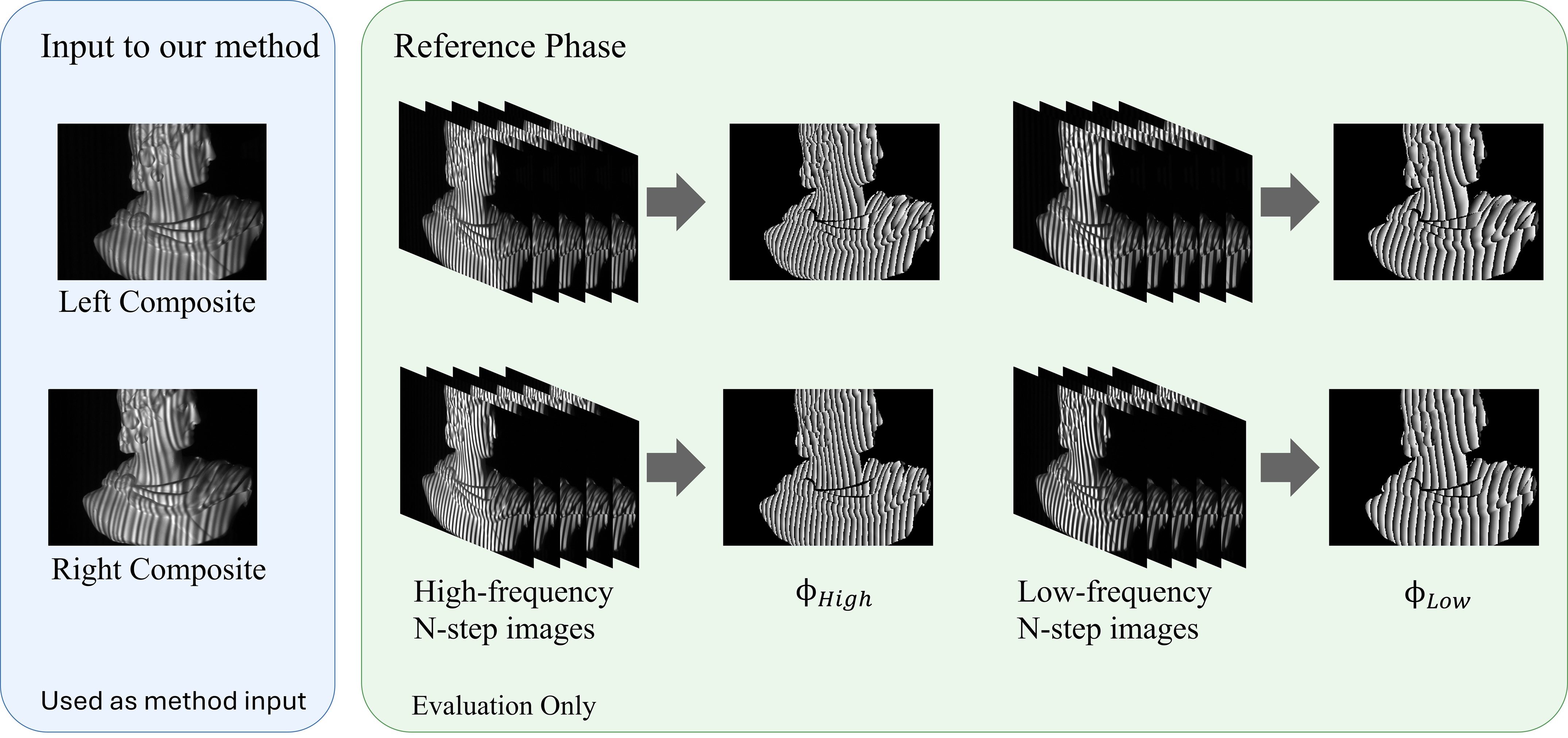}
    \caption{
    Overview of the dataset composition and reference generation.
    }
    \label{fig:dataset}
\end{figure}

\subsection{Dataset and Evaluation Metric}
\noindent\textbf{Dataset.} Fig.~\ref{fig:dataset} illustrates one sample from our real-world dataset. Each sample contains a pair of left and right composite fringe images, which are used as the inputs to our method. For evaluation, we additionally capture low- and high-frequency $N$-step sinusoidal patterns for each view and compute the corresponding reference wrapped phases using phase shifting. 
The wrapped phases are stored as cosine and sine components. 
In our acquisition setup, we use 9 phase-shifted images for the high-frequency pattern and 14 phase-shifted images for the low-frequency pattern. 
Gray-coded patterns are also captured to obtain the ground-truth 3D geometry. 
These additional sinusoidal and Gray-coded patterns are used only for reference generation and evaluation, not as inputs to the proposed method. 
The dataset contains 1,200 samples in total, split into 1,000 training samples, 100 validation samples, and 100 test samples. 
The splits are object-disjoint, meaning that no object instance is shared across the training, validation, and test sets.

\vspace{1.0em}
\noindent\textbf{Evaluation metrics.} 
Our 3D reconstruction framework does not utilize unwrapped phase in correspondence search. Instead, low- and high-frequency wrapped phases are used for phase-based candidate reduction stereo matching. However, wrapped phase is periodic, so a small wrapped phase error alone does not demonstrate whether the correct fringe order has been recovered. Therefore, to evaluate phase correctness at the fringe-order level, we construct a pseudo unwrapped phase only for evaluation. This pseudo unwrapped phase is computed after 3D reconstruction. It is not used for correspondence search or 3D reconstruction.

Given the reconstructed 3D point $\mathbf{x}$, we reproject it into the projector phase coordinate using the calibrated projector-camera geometry:
\begin{equation}
    \hat{\Phi}_{geo}(\mathbf{x})
    =
    f_{\mathrm{proj}}
    \left(
        \mathbf{x}; \Theta_{\mathrm{proj}}
    \right),
\end{equation}
where $f_{\mathrm{proj}}(\cdot)$ denotes the calibrated projection from a 3D point to the projector phase coordinate, and $\Theta_{\mathrm{proj}}$ denotes the projector calibration parameters.

The pseudo unwrapped phase is then obtained by
\begin{equation}
    \hat{\Phi}^{pseudo}_{h}(\mathbf{x})
    =
    \hat{\phi}_{h}(\mathbf{x})
    +
    2\pi \hat{k}_{h}(\mathbf{x}).
\end{equation}

The predicted high-frequency fringe order is estimated as
\begin{equation}
    \hat{k}_{h}(\mathbf{x})
    =
    \mathrm{round}
    \left(
        \frac{
            \hat{\Phi}_{geo}(\mathbf{x}) - \hat{\phi}_{h}(\mathbf{x})
        }{2\pi}
    \right),
\end{equation}
where $\hat{\phi}_{h}(\mathbf{x})$ is the predicted high-frequency wrapped phase.

The reference high-frequency fringe order is computed from the reference wrapped and unwrapped phases:
\begin{equation}
    k^{gt}_{h}(\mathbf{x})
    =
    \mathrm{round}
    \left(
        \frac{
            \Phi^{gt}_{h}(\mathbf{x}) - \phi^{gt}_{h}(\mathbf{x})
        }{2\pi}
    \right),
\end{equation}
where $\phi^{gt}_{h}(\mathbf{x})$ and $\Phi^{gt}_{h}(\mathbf{x})$ denote the reference high-frequency wrapped and unwrapped phases, respectively.

For evaluation, we define the GT-valid foreground region as the pixels where the reference depth $z^{gt}$ is valid. 
For each method, valid pixels are defined as the pixels where both the reconstructed depth $\hat{z}$ and the reference depth $z^{gt}$ are valid. 
The valid-pixel ratio is computed as the ratio of method-valid pixels to GT-valid foreground pixels.

The fringe-order accuracy is defined as
\begin{equation}
    \mathrm{FOAcc}
    =
    \frac{1}{N}
    \sum_{\mathbf{x}}
    \mathbb{1}
    \left[
        \hat{k}_{h}(\mathbf{x}) = k^{gt}_{h}(\mathbf{x})
    \right],
\end{equation}
where $N$ denotes the number of GT-valid foreground pixels. 
Pixels without valid reconstruction are counted as incorrect.

We then compare the pseudo unwrapped phase $\hat{\Phi}^{pseudo}_{h}$ with the reference unwrapped phase $\Phi^{gt}_{h}$ and report the mean absolute error and root mean squared error, denoted as MAE$_{\Phi}$ and RMSE$_{\Phi}$, respectively. 
Since this metric evaluates the unwrapped phase coordinate, the phase difference is not wrapped back into $[-\pi,\pi]$. MAE$_{\Phi}$ and RMSE$_{\Phi}$ are computed over the valid pixels of each method.

For geometric evaluation, we compare the reconstructed depth $\hat{z}$ with the reference depth $z^{gt}$ and report MAE$_z$ and RMSE$_z$, which are computed over the valid pixels of each method. 
In addition, we report threshold-based depth accuracy, denoted as Acc.(1mm) and Acc.(3mm). 
Acc.($\tau$) is defined as the percentage of GT-valid foreground pixels whose reconstructed depth is valid and whose depth error is smaller than $\tau$ mm. 
Thus, invalid reconstructions are counted as failures in the threshold-based accuracy. We additionally report the 95th-percentile absolute depth error, denoted as P95 AE$_z$, computed over the valid pixels of each method. 
For complementary analysis, we also report common-valid-mask MAE$_z$ and RMSE$_z$, where all methods are evaluated over the intersection of their valid pixel regions.

\subsection{Comparison with Baseline Methods}

We compare the proposed method with representative transform-based baselines, including WFT, Wavelet-1D, Wavelet-2D, and Shearlet transforms. All baseline methods use the same single-shot composite pattern as input and estimate the high- and low-frequency wrapped phases. For a fair comparison, the parameters of each transform-based method were selected on the validation set and fixed for all test samples. No sample-specific parameter tuning was performed during evaluation. All methods, including the proposed method and the baselines are evaluated using the same 3D reconstruction pipeline. Corresponding pairs are determined by phase-based stereo matching using the estimated dual-frequency wrapped phases. For reference generation, we computed the wrapped phase from 9 phase-shifted fringe images using an N-step phase-shifting method and obtained the unwrapped phase using 7 Gray-code patterns. The unwrapped phase was used only to generate the ground-truth correspondence and geometry, not as input to the evaluated single-shot methods.

\begin{table}[t]
\centering
\resizebox{\linewidth}{!}{
\begin{tabular}{lcccccccc}
\toprule
Method 
& Valid (\%) $\uparrow$
& FOAcc (\%) $\uparrow$
& MAE$_{\Phi}$ $\downarrow$
& RMSE$_{\Phi}$ $\downarrow$
& MAE$_z$ $\downarrow$
& RMSE$_z$ $\downarrow$
& Acc.(1mm) $\uparrow$
& Acc.(3mm) $\uparrow$ \\
\midrule
WFT 
& 86.07
& 80.14
& 0.824
& 3.998
& 0.933
& 4.591
& 77.66
& 84.38 \\

Wavelet-1D 
& 85.97
& 80.69
& 0.507
& 2.590
& 0.420
& 2.794
& 82.90
& 85.51 \\

Wavelet-2D 
& 84.75
& 79.94
& 0.492
& 2.552
& 0.402
& 2.785
& 81.84
& 84.29 \\

Shearlet 
& 86.89
& 80.05
& 0.726
& 3.131
& 0.660
& 3.550
& 80.84
& 85.93 \\

Ours 
& \textbf{95.07}
& \textbf{91.61}
& \textbf{0.298}
& \textbf{2.120}
& \textbf{0.367}
& \textbf{1.804}
& \textbf{90.06}
& \textbf{94.16} \\
\bottomrule
\end{tabular}
}
\caption{Quantitative comparison of phase separation and depth reconstruction.
Valid, FOAcc, and Acc. are normalized by the GT-valid foreground region, while MAE/RMSE metrics are computed over each method's valid pixels.}
\label{tab:quantitative_results}
\end{table}

Table~\ref{tab:quantitative_results} reports the quantitative comparison of phase separation and 3D reconstruction. The proposed method achieves the highest valid-pixel ratio of 95.07\%, indicating that it reconstructs a larger portion of the GT-valid foreground region. It also obtains the highest FOAcc of 91.61\%, demonstrating more reliable fringe-order recovery and robust phase-ambiguity resolution. The pseudo unwrapped phase errors are reduced to the lowest MAE$_{\Phi}$ and RMSE$_{\Phi}$ of 0.298 rad and 2.120 rad, respectively. In terms of depth reconstruction, the proposed method achieves MAE$_z$ and RMSE$_z$ of 0.367 mm and 1.804 mm, respectively, outperforming the strongest transform-based baseline, Wavelet-2D, which obtains 0.402 mm and 2.785 mm. The proposed method also obtains the highest threshold-based depth accuracies, with Acc.(1mm) and Acc.(3mm) of 90.06\% and 94.16\%, respectively. These results indicate that the improved phase separation quality translates into better 3D reconstruction performance.
It is worth noting that RMSE$_z$ remains substantially larger than MAE$_z$ for all methods, including the proposed one (1.804 mm versus 0.367 mm).
This gap indicates that the overall error distribution is dominated by a small number of large-error outliers rather than by widespread inaccuracy, since RMSE is more sensitive to such outliers than MAE.
These outliers mainly arise near object boundaries and depth discontinuities, where fringe-order errors and triangulation instability occur, which is consistent with the residual boundary errors discussed in the qualitative results and the conclusion.

\begin{table}[t]
\centering
\resizebox{\linewidth}{!}{
\begin{tabular}{lccc}
\toprule
Method
& Common-mask MAE$_z$ $\downarrow$
& Common-mask RMSE$_z$ $\downarrow$
& P95 AE$_z$ $\downarrow$ \\
\midrule
WFT 
& 0.574
& 3.121
& 2.299 \\

Wavelet-1D 
& 0.254
& 1.541
& \textbf{0.790} \\

Wavelet-2D 
& 0.260
& 1.745
& 0.772 \\

Shearlet 
& 0.400
& 2.117
& 1.140 \\

Ours 
& \textbf{0.176}
& \textbf{0.507}
& 0.968 \\
\bottomrule
\end{tabular}
}
\caption{Additional analysis of depth errors using common valid regions and percentile-based evaluation.
Common-mask MAE$_z$ and RMSE$_z$ are computed over the intersection of valid pixel regions across all methods, while P95 AE$_z$ is computed over each method's valid pixels.}
\label{tab:additional_depth_analysis}
\end{table}

Table~\ref{tab:additional_depth_analysis} provides a complementary analysis of depth errors to examine the effect of different valid-pixel regions. The common-mask evaluation compares all methods over the same valid pixel set. The proposed method achieves the lowest common-mask MAE$_z$ and RMSE$_z$ of 0.176 mm and 0.507 mm, respectively. This indicates that the improvement is not solely due to its broader valid-pixel coverage, but is also observed on the commonly valid pixels. Unlike the common-mask metrics, P95 AE$_z$ is computed over each method's valid pixels. Wavelet-1D shows a lower P95 AE$_z$ of 0.790 mm than the proposed method, which obtains 0.968 mm. This suggests that Wavelet-1D is locally accurate within its valid region, but its lower P95 should be interpreted together with its lower valid-pixel ratio.

\begin{figure*}[!htbp]
    \centering
    \includegraphics[width=\textwidth]{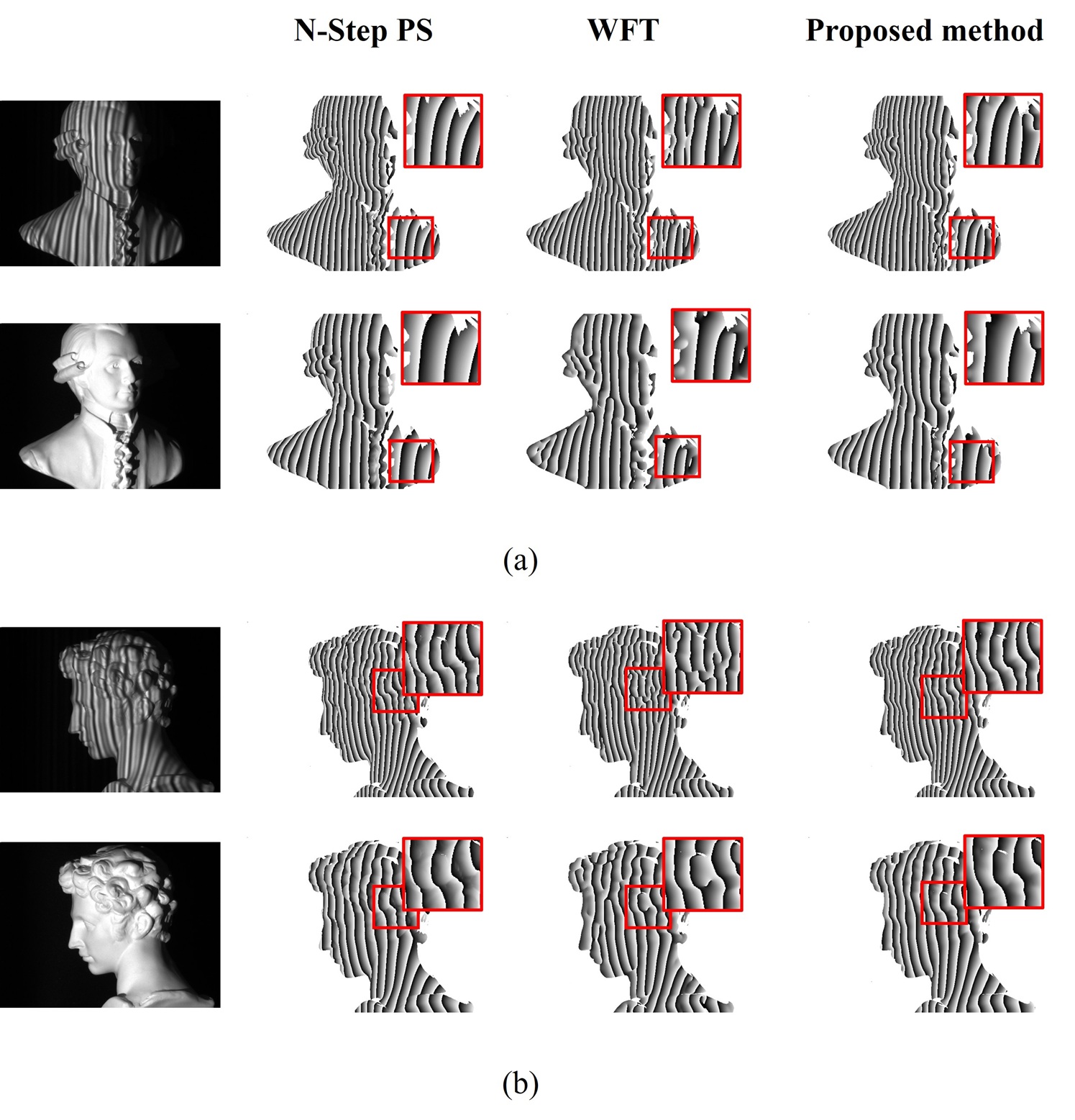}
    \caption{
    Qualitative comparison of high- and low-frequency wrapped phase maps on two representative samples.
    The rightmost column shows the captured images under composite-fringe and white-light illumination.
    The remaining columns show the phase maps obtained by N-step phase shifting, WFT, and the proposed method, respectively.
    The upper and lower rows correspond to the high- and low-frequency components.
    }
    \label{fig:qualitative_phase}
\end{figure*}

Fig.~\ref{fig:qualitative_phase} presents qualitative high- and low-frequency wrapped phase separation results. WFT is shown as a representative transform-based baseline. While WFT captures the global fringe structure, local phase artifacts remain in complex texture regions and near object boundaries. The proposed method produces phase maps that are more consistent with the N-step reference, which supports the improved fringe-order accuracy reported in Table~\ref{tab:quantitative_results}.

\begin{figure*}[!htbp]
    \centering
    \includegraphics[width=\textwidth]{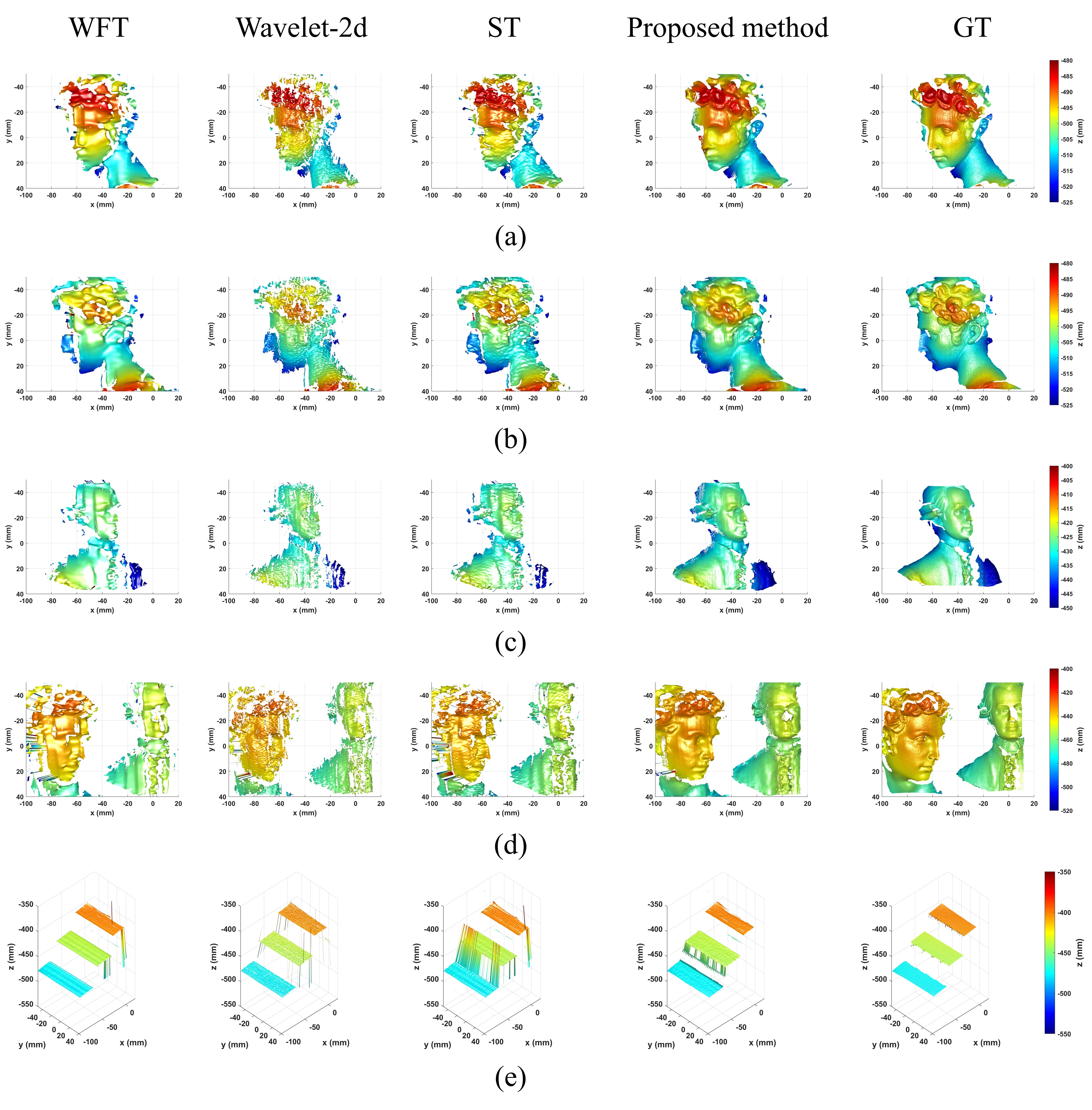}
    \caption{
    Qualitative comparison of reconstructed 3D results.
    Rows (a)--(c) show three different views of the statue sample, while rows (d) and (e) show the isolated object and stair samples, respectively.
    }
    \label{fig:qualitative_3d}
\end{figure*}

Fig.~\ref{fig:qualitative_3d} shows qualitative 3D reconstruction results. The transform-based baselines exhibit missing regions and unstable surfaces, whereas the proposed method reconstructs more complete surfaces. The proposed method maintains stable reconstruction across different scenes, including the statue, isolated-object, and stair samples. This trend is consistent with the higher valid-pixel ratio and threshold-based depth accuracy reported in Table~\ref{tab:quantitative_results}. Overall, the experimental results demonstrate reliable single-shot reconstruction performance using only a single composite fringe image.

\subsection{Ablation Study}

\begin{table}[t]
\centering
\resizebox{\linewidth}{!}{
\begin{tabular}{lcccccccc}
\toprule
Configuration
& Valid (\%) $\uparrow$
& FOAcc (\%) $\uparrow$
& MAE$_{\Phi}$ $\downarrow$
& RMSE$_{\Phi}$ $\downarrow$
& MAE$_z$ $\downarrow$
& RMSE$_z$ $\downarrow$
& Acc.(1mm) $\uparrow$
& Acc.(3mm) $\uparrow$ \\
\midrule
A0
& 93.33
& 88.01
& 0.559
& 3.111
& 0.725
& 3.146
& 82.02
& 91.36 \\

A1
& 93.63
& 88.26
& 0.521
& 2.908
& 0.651
& 2.714
& 82.70
& 91.76 \\

A2
& 93.87
& 89.38
& 0.457
& 2.791
& 0.543
& 2.555
& 86.49
& 92.51 \\

A3
& \textbf{95.07}
& \textbf{91.61}
& \textbf{0.298}
& \textbf{2.120}
& \textbf{0.367}
& \textbf{1.804}
& \textbf{90.06}
& \textbf{94.16} \\
\bottomrule
\end{tabular}
}
\caption{Ablation study on the proposed loss terms. The loss terms are added cumulatively from A0 to A3.}
\label{tab:ablation_study}
\end{table}

We conduct an ablation study to analyze the contribution of each loss term.
A0 denotes the minimal baseline configuration, including the image reconstruction loss in Eq.~\eqref{eq:loss_rec} and phase consistency losses in Eq.~\eqref{eq:loss_scale} and Eq.~\eqref{eq:loss_dir}. A1 extends A0 by adding the phase regularization term in Eq.~\eqref{eq:loss_cont}. A2 further extends A1 with the modulation regularization term in Eq.~\eqref{eq:loss_mod_sigma_ratio}.
A3 denotes the full model, which further includes the shared edge consistency loss in Eq.~\eqref{eq:loss_edge}. The loss terms are added cumulatively from A0 to A3.

\begin{figure*}[!htbp]
    \centering
    \includegraphics[width=\textwidth]{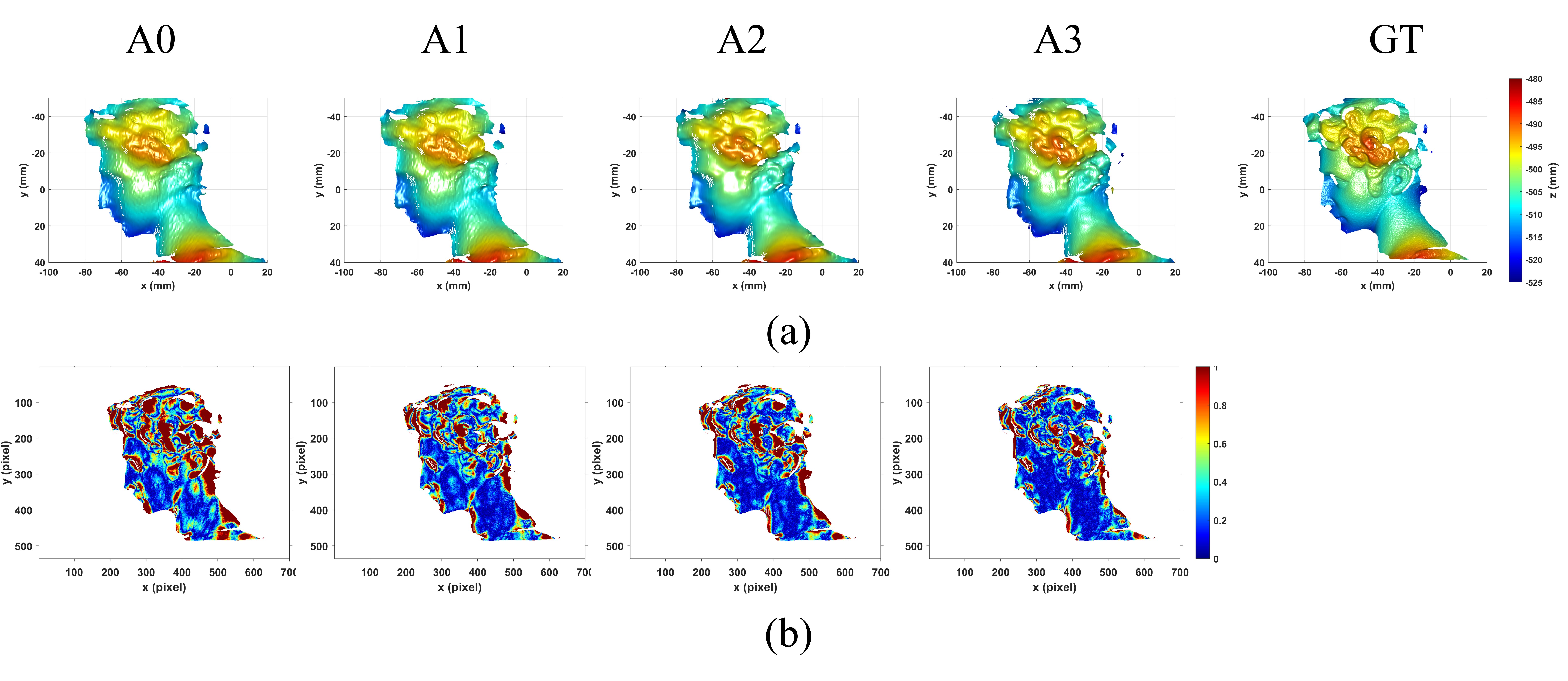}
    \caption{
    Qualitative ablation results for 3D reconstruction.
    Row (a) shows the reconstructed 3D results, and row (b) shows the corresponding error maps.
    The column descriptions are provided in the figure.
    }
    \label{fig:ablation_3d}
\end{figure*}

Table~\ref{tab:ablation_study} reports the quantitative ablation results.
The performance improves progressively from A0 to A3 as the proposed loss terms are added. Fig.~\ref{fig:ablation_3d} shows the qualitative ablation results, including 3D reconstruction results and the corresponding error maps. As the loss terms are progressively added, large-error regions are gradually reduced and the reconstructed surfaces become more stable. These results indicate that the proposed loss terms are complementary and jointly contribute to reliable phase separation and accurate 3D reconstruction.

\section{Conclusion}

In this work, we proposed a single-shot self-supervised phase refinement framework for composite fringe patterns.
The proposed framework is trained without GT phase or depth supervision and exploits structural constraints based on the scale and direction relationships between the gradients of low- and high-frequency phases.
In addition, we introduced a soft edge consistency loss that aligns edge structures among phase variations, modulation amplitudes, and texture, thereby helping preserve object boundaries and fine geometric structures.

Experimental results demonstrated that the proposed method improves the valid-pixel ratio, FOAcc, and depth accuracy compared with representative transform-based baselines.
These results indicate that the proposed method provides reliable dual-frequency phase cues, leading to more complete and accurate 3D reconstruction.
Despite these improvements, a performance gap remains compared with fully supervised methods, and residual errors still appear near object boundaries and depth discontinuities.
Future work will focus on improving edge-region reconstruction and reducing the performance gap with supervised methods through stronger boundary-aware refinement and self-supervised constraints.

\section*{Acknowledgements}
This research was supported by the Technology Innovation Program (Project Name: Development of AI autonomous continuous production system technology for gas turbine blade maintenance and regeneration for power generation, Project Number: RS-2025-25447257, Contribution Rate: 30\%) funded by the Ministry of Trade, Industry and Resources (MOTIR, Korea), the Culture, Sports and Tourism R\&D Program through the Korea Creative Content Agency grant funded by the Ministry of Culture, Sports and Tourism in 2024 (Project Name: Global Talent for Generative AI Copyright Infringement and Copyright Theft, Project Number: RS-2024-00398413, Contribution Rate: 50\%), the National Research Foundation of Korea (NRF) grant funded by the Korea government (MSIT) (Project Number: RS-2026-25498577, Contribution Rate: 15\%), and the Yonsei Institute for Embodied Intelligence (Contribution Rate: 5\%).





\bibliographystyle{elsarticle-num} 
\bibliography{cas-refs}

@article{fu2024high,
  title={High-accuracy fringe projection profilometry without phase unwrapping based on multi-view geometry constraints},
  author={Fu, Yanjun and Luo, Lin and Zhong, Kejun and Li, Fangfang},
  journal={Optics Express},
  volume={32},
  number={22},
  pages={38449--38464},
  year={2024},
  publisher={Optica Publishing Group}
}

@article{tang2024dual,
  title={Dual frequency composite pattern temporal phase unwrapping for 3D surface measurement},
  author={Tang, Tao and Zhang, Yu and Wan, Yingying and Peng, Jianping and Li, Jinlong and Luo, Lin},
  journal={Scientific Reports},
  volume={14},
  number={1},
  pages={24992},
  year={2024},
  publisher={Nature Publishing Group UK London}
}

@article{zhang2006novel,
  title={Novel method for structured light system calibration},
  author={Zhang, Song and Huang, Peisen S},
  journal={Optical Engineering},
  volume={45},
  number={8},
  pages={083601--083601},
  year={2006},
  publisher={Society of Photo-Optical Instrumentation Engineers}
}

@article{gorthi2010fringe,
  title={Fringe projection techniques: whither we are?},
  author={Gorthi, Sai Siva and Rastogi, Pramod},
  year={2010},
  publisher={Elsevier}
}

@article{zuo2018phase,
  title={Phase shifting algorithms for fringe projection profilometry: A review},
  author={Zuo, Chao and Feng, Shijie and Huang, Lei and Tao, Tianyang and Yin, Wei and Chen, Qian},
  journal={Optics and lasers in engineering},
  volume={109},
  pages={23--59},
  year={2018},
  publisher={Elsevier}
}

@article{zhang2018absolute,
  title={Absolute phase retrieval methods for digital fringe projection profilometry: A review},
  author={Zhang, Song},
  journal={Optics and lasers in engineering},
  volume={107},
  pages={28--37},
  year={2018},
  publisher={Elsevier}
}

@article{feng2021calibration,
  title={Calibration of fringe projection profilometry: A comparative review},
  author={Feng, Shijie and Zuo, Chao and Zhang, Liang and Tao, Tianyang and Hu, Yan and Yin, Wei and Qian, Jiaming and Chen, Qian},
  journal={Optics and lasers in engineering},
  volume={143},
  pages={106622},
  year={2021},
  publisher={Elsevier}
}

@article{feng2018high,
  title={High dynamic range 3D measurements with fringe projection profilometry: a review},
  author={Feng, Shijie and Zhang, Liang and Zuo, Chao and Tao, Tianyang and Chen, Qian and Gu, Guohua},
  journal={Measurement Science and Technology},
  volume={29},
  number={12},
  pages={122001},
  year={2018},
  publisher={IOP Publishing}
}

@article{zhang2010recent,
  title={Recent progresses on real-time 3D shape measurement using digital fringe projection techniques},
  author={Zhang, Song},
  journal={Optics and lasers in engineering},
  volume={48},
  number={2},
  pages={149--158},
  year={2010},
  publisher={Elsevier}
}

@article{zhang2018high,
  title={High-speed 3D shape measurement with structured light methods: A review},
  author={Zhang, Song},
  journal={Optics and lasers in engineering},
  volume={106},
  pages={119--131},
  year={2018},
  publisher={Elsevier}
}

@article{liu2024deep,
  title={Deep learning in fringe projection: A review},
  author={Liu, Haoyue and Yan, Ning and Shao, Bofan and Yuan, Shuaipeng and Zhang, Xiaodong},
  journal={Neurocomputing},
  volume={581},
  pages={127493},
  year={2024},
  publisher={Elsevier}
}

@article{wang2025single,
  title={Single-shot super-resolved fringe projection profilometry (SSSR-FPP): 100,000 frames-per-second 3D imaging with deep learning},
  author={Wang, Bowen and Chen, Wenwu and Qian, Jiaming and Feng, Shijie and Chen, Qian and Zuo, Chao},
  journal={Light: Science \& Applications},
  volume={14},
  number={1},
  pages={70},
  year={2025},
  publisher={Nature Publishing Group UK London}
}

@article{huang2010comparison,
  title={Comparison of Fourier transform, windowed Fourier transform, and wavelet transform methods for phase extraction from a single fringe pattern in fringe projection profilometry},
  author={Huang, Lei and Kemao, Qian and Pan, Bing and Asundi, Anand Krishna},
  journal={Optics and Lasers in Engineering},
  volume={48},
  number={2},
  pages={141--148},
  year={2010},
  publisher={Elsevier}
}

@article{kemao2007two,
  title={Two-dimensional windowed Fourier transform for fringe pattern analysis: principles, applications and implementations},
  author={Kemao, Qian},
  journal={Optics and Lasers in Engineering},
  volume={45},
  number={2},
  pages={304--317},
  year={2007},
  publisher={Elsevier}
}

@article{li2009spatial,
  title={Spatial carrier fringe pattern phase demodulation by use of a two-dimensional real wavelet},
  author={Li, Sikun and Su, Xianyu and Chen, Wenjing},
  journal={Applied optics},
  volume={48},
  number={36},
  pages={6893--6906},
  year={2009},
  publisher={Optical Society of America}
}

@article{gao2024one,
  title={One-shot structured light illumination based on shearlet transform},
  author={Gao, Rui and Zhao, Xiaobing and Lau, Daniel L and Zhang, Bo and Xu, Bin and Liu, Kai},
  journal={Optics Express},
  volume={32},
  number={17},
  pages={30182--30198},
  year={2024},
  publisher={Optica Publishing Group}
}

@article{li2022composite,
  title={Composite fringe projection deep learning profilometry for single-shot absolute 3D shape measurement},
  author={Li, Yixuan and Qian, Jiaming and Feng, Shijie and Chen, Qian and Zuo, Chao},
  journal={Optics express},
  volume={30},
  number={3},
  pages={3424--3442},
  year={2022},
  publisher={Optica Publishing Group}
}

@article{yue2007fourier,
  title={Fourier transform profilometry based on composite structured light pattern},
  author={Yue, Hui-Min and Su, Xian-Yu and Liu, Yong-Zhi},
  journal={Optics \& Laser Technology},
  volume={39},
  number={6},
  pages={1170--1175},
  year={2007},
  publisher={Elsevier}
}

@article{li2025deep,
  title={Deep-learning-enabled dual-frequency composite fringe projection profilometry for single-shot absolute 3D shape measurement},
  author={Li, Yixuan and Qian, Jiaming and Feng, Shijie and Chen, Qian and Zuo, Chao},
  journal={Opto-Electronic Advances},
  volume={5},
  number={5},
  pages={210021--1},
  year={2025}
}

@article{wang2021single,
  title={Single-shot fringe projection profilometry based on deep learning and computer graphics},
  author={Wang, Fanzhou and Wang, Chenxing and Guan, Qingze},
  journal={Optics Express},
  volume={29},
  number={6},
  pages={8024--8040},
  year={2021},
  publisher={Optical Society of America}
}

@article{jiang2024deep,
  title={Deep-learning-based single-shot fringe projection profilometry using spatial composite pattern},
  author={Jiang, Yansong and Qin, Jiayi and Liu, Yuankun and Yang, Menglong and Cao, Yiping},
  journal={IEEE Transactions on Instrumentation and Measurement},
  volume={73},
  pages={1--14},
  year={2024},
  publisher={IEEE}
}

@article{fu2024deep,
  title={Deep learning-based binocular composite color fringe projection profilometry for fast 3D measurements},
  author={Fu, Yanjun and Huang, Yiliang and Xiao, Wei and Li, Fangfang and Li, Yunzhan and Zuo, Pengfei},
  journal={Optics and Lasers in Engineering},
  volume={172},
  pages={107866},
  year={2024},
  publisher={Elsevier}
}

@article{wu2024depth,
  title={Depth acquisition from dual-frequency fringes based on end-to-end learning},
  author={Wu, Yingchun and Wang, Zihao and Liu, Li and Yang, Na and Zhao, Xianling and Wang, Anhong},
  journal={Measurement Science and Technology},
  volume={35},
  number={4},
  pages={045203},
  year={2024},
  publisher={IOP Publishing}
}

@article{chen2024deep,
  title={Deep learning-based frequency-multiplexing composite-fringe projection profilometry technique for one-shot 3D shape measurement},
  author={Chen, Yifei and Kang, Jiehu and Feng, Luyuan and Yuan, Leiwen and Liang, Jian and Zhao, Zongyang and Wu, Bin},
  journal={Measurement},
  volume={233},
  pages={114640},
  year={2024},
  publisher={Elsevier}
}

@article{wang2023deep,
  title={Deep learning-based end-to-end 3D depth recovery from a single-frame fringe pattern with the MSUNet++ network},
  author={Wang, Chao and Zhou, Pei and Zhu, Jiangping},
  journal={Optics Express},
  volume={31},
  number={20},
  pages={33287--33298},
  year={2023},
  publisher={Optica Publishing Group}
}

@article{qin2025single,
  title={Single-shot phase-shifting composition fringe projection profilometry by multi-attention fringe restoration network},
  author={Qin, Jiayi and Jiang, Yansong and Cao, Yiping and Wu, Haitao},
  journal={Neurocomputing},
  volume={634},
  pages={129908},
  year={2025},
  publisher={Elsevier}
}

@article{wang2025end,
  title={End-to-end single-shot composite color FPP network for multiple separated objects reconstruction},
  author={Wang, Lianpo and Xing, Yanyang},
  journal={Measurement},
  volume={246},
  pages={116697},
  year={2025},
  publisher={Elsevier}
}

@article{yu2023untrained,
  title={Untrained deep learning-based phase retrieval for fringe projection profilometry},
  author={Yu, Haotian and Chen, Xiaoyu and Huang, Ruobing and Bai, Lianfa and Zheng, Dongliang and Han, Jing},
  journal={Optics and Lasers in Engineering},
  volume={164},
  pages={107483},
  year={2023},
  publisher={Elsevier}
}

@article{yu2022untrained,
  title={Untrained deep learning-based fringe projection profilometry},
  author={Yu, Haotian and Han, Bowen and Bai, Lianfa and Zheng, Dongliang and Han, Jing},
  journal={APL photonics},
  volume={7},
  number={1},
  year={2022},
  publisher={AIP Publishing}
}

@article{fan2021unsupervised,
  title={Unsupervised deep learning for 3D reconstruction with dual-frequency fringe projection profilometry},
  author={Fan, Sizhe and Liu, Shaoli and Zhang, Xu and Huang, Hao and Liu, Wei and Jin, Peng},
  journal={Optics express},
  volume={29},
  number={20},
  pages={32547--32567},
  year={2021},
  publisher={Optical Society of America}
}

@article{tan2024weakly,
  title={Weakly supervised depth estimation for 3d imaging with single camera fringe projection profilometry},
  author={Tan, Chunqian and Song, Wanzhong},
  journal={Sensors},
  volume={24},
  number={5},
  pages={1701},
  year={2024},
  publisher={MDPI}
}

@article{gao2024weakly,
  title={Weakly supervised phase unwrapping for single-camera fringe projection profilometry},
  author={Gao, Xiaoming and Song, Wanzhong},
  journal={Optics Communications},
  volume={557},
  pages={130308},
  year={2024},
  publisher={Elsevier}
}

@article{wang2026end,
  title={End-to-end single-shot fringe projection profilometry based on semi-supervised learning},
  author={Wang, Huitao and Wang, Lianpo},
  journal={Measurement},
  pages={121076},
  year={2026},
  publisher={Elsevier}
}

@inproceedings{chen2023self,
  title={Self-supervised monocular depth estimation: Solving the edge-fattening problem},
  author={Chen, Xingyu and Zhang, Ruonan and Jiang, Ji and Wang, Yan and Li, Ge and Li, Thomas H},
  booktitle={Proceedings of the IEEE/CVF Winter Conference on Applications of Computer Vision},
  pages={5776--5786},
  year={2023}
}





\end{document}